\documentclass[10pt,twocolumn,letterpaper]{article}
\usepackage{cvpr}
\usepackage{xcolor}
\usepackage{times}
\usepackage{pgffor}
\usepackage{overpic}
\usepackage{graphicx}
\usepackage{amsmath}
\usepackage{amssymb}
\usepackage{xspace}
\usepackage{enumitem}
\usepackage{tabularx}
\usepackage{multirow}
\usepackage[numbers,square,sort&compress]{natbib}
\usepackage[
  pagebackref=true,
  breaklinks=true,
  letterpaper=true,
  colorlinks,
  bookmarks=false]{hyperref}

\cvprfinalcopy
\def\cvprPaperID{17}

\setitemize{noitemsep,topsep=0pt,parsep=0pt,partopsep=0pt,nolistsep,leftmargin=*}

\newcommand{\dataseturl}{\url{http://www.robots.ox.ac.uk/~vgg/data/fgvc-aircraft/}}

\begin{document}
\title{Fine-Grained Visual Classification of Aircraft}
\author{
Subhransu Maji  \\ 
TTI Chicago \\ 
\small{\textsf{smaji@ttic.edu}} \and
Esa Rahtu\ \ \ \ \ \  Juho Kannala \\ 
University of Oulu, Finland \\ 
\small{\textsf{\{erahtu, jkannala\}@ee.oulu.fi}}\and
Matthew Blaschko \\ 
\'Ecole Centrale Paris \\ 
\small{\textsf{matthew.blaschko@ecp.fr}}\and
Andrea Vedaldi \\ 
University of Oxford \\ 
\small{\textsf{vedaldi@robots.ox.ac.uk}}
}
\maketitle
% --------------------------------------------------------------------

\begin{figure*}
\includegraphics[width=\linewidth]{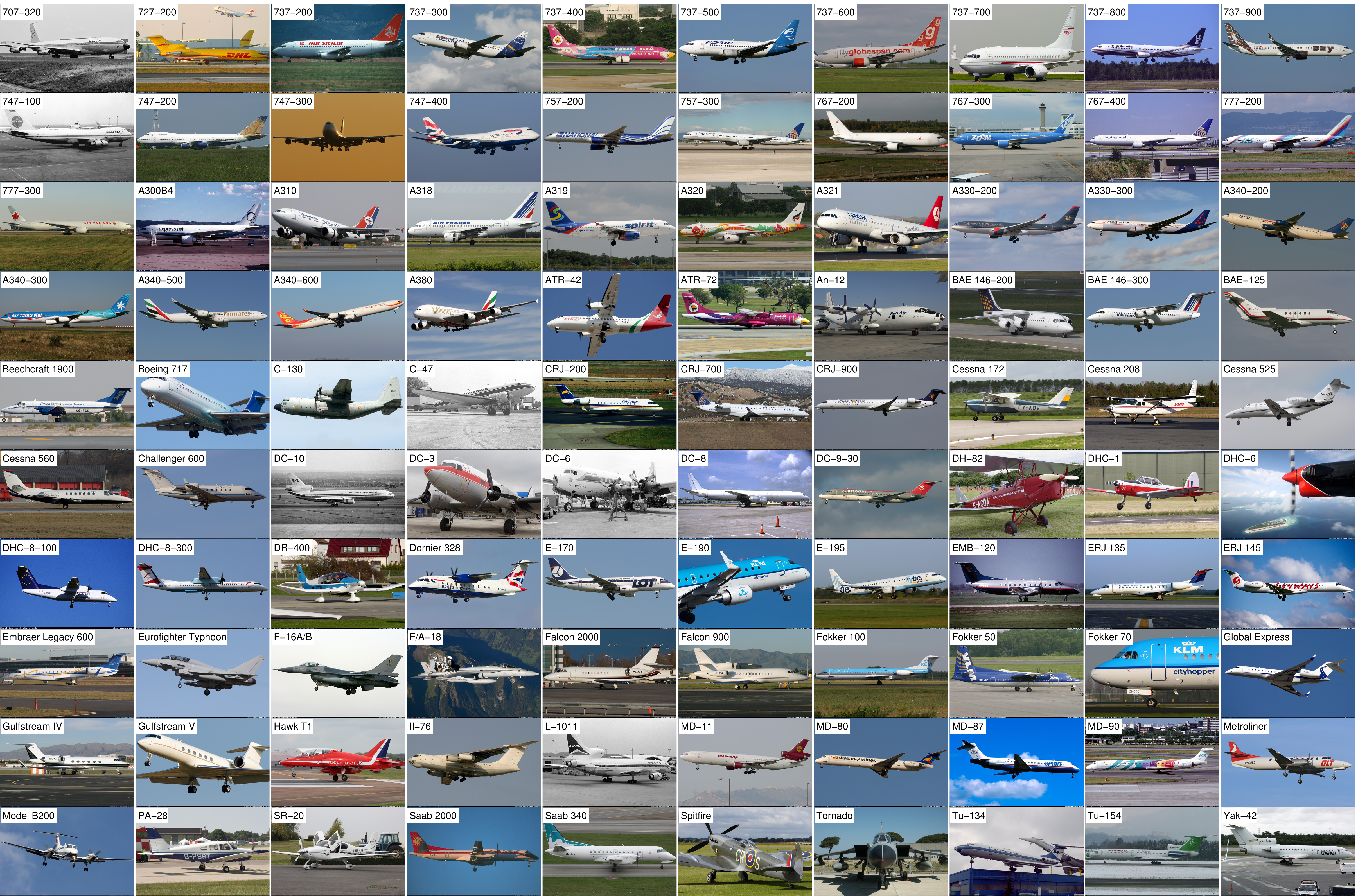}
\caption{Our dataset contains 100 variants of aircrafts shown above. These are also annotated with their family and manufacturer, as well as bounding boxes.}\label{f:variants}
\end{figure*}

\begin{abstract}
This paper introduces {\em FGVC-Aircraft}, a new dataset containing 10,000 images of aircraft spanning 100 aircraft models, organised in a three-level hierarchy. At the finer level, differences between models are often subtle but always visually measurable, making visual recognition challenging but possible. A benchmark is obtained by defining corresponding classification tasks and evaluation protocols, and baseline results are presented. The construction of this dataset was made possible by the work of aircraft enthusiasts, a strategy that can extend to the study of number of other object classes. Compared to the domains usually considered in fine-grained visual classification (FGVC), for example animals, aircraft are rigid and hence less deformable. They, however, present other interesting modes of variation, including purpose, size, designation, structure, historical style, and branding. 
\end{abstract}

% --------------------------------------------------------------------
\section{Introduction}\label{s:intro}
% --------------------------------------------------------------------

%We argue that aircraft recognition may be of particular interest to the community due to the huge variety of aircraft designs, characterised both by substantial structural differences (e.g. different number of wings) as well as minor variations (e.g. revisions of a blueprint). 

In this paper, we introduce {\em FGVC-Aircraft}, a novel dataset aimed at studying the problem of fine-grained recognition of aircraft models (Fig.~\ref{f:variants}, Sect.~\ref{s:data}). The new data includes 10,000 airplane images spanning 100 different models, organised in a hierarchical manner. All models are visually distinguishable, even though in many cases the differences are  subtle, making classification challenging and interesting.

Airplanes are an alternative to objects typically considered in {\em fine-grained visual classification} (FGVC) such as birds~\cite{wah11the-caltech-ucsd} and pets~\cite{parkhi12cats,liu12dog-breed,khosla11novel}. Compared to these domains, aircraft classification has several interesting aspects. First, aircraft designs vary significantly depending on the plane size (from home-built to large carriers), designation (private, civil, military), purpose (transporter, carrier, training, sport, fighter, etc.), and technological factors such as propulsion (glider, propellor, jet). Overall, thousands of different airplane models exist or have existed.  An interesting mode of variation, which is is not shared with categories such as animals, is the fact that the {\em structure} of aircraft can change with their design. For example, the number of wings, undercarriages, wheels per undercarriage, engines, etc. varies. Second, the aircraft designs exhibit systematic {\em historical variations} in their style. Thirdly, the same aircraft models are used by different airliner companies, resulting in variable livery branding. Finally, aircraft are largely rigid objects, reducing the impact of deformability on classification performance, and allowing one to focus on the other aspects of FGVC.

Our contributions are three-fold: (i) we introduce a new large dataset of aircraft images with detailed model annotations; (ii) we describe how this data was collected using on-line resources and the work of hobbyists and enthusiasts -- a method that may be applicable to other object classes; and (iii) we present baseline results on aircraft model identification. Sect.~\ref{s:data} describes the content of {\em FGVC-Aircraft}, including task definitions and evaluation protocols, Sect.~\ref{s:construction} the dataset construction, Sect.~\ref{s:baselines} examines the performance of a baseline classifier on the data, and Sect.~\ref{s:summary} summarises the contributions, giving further details on the data usage policy.

\begin{figure*}[!t]
\centering
\begin{tabular}{cc}
\includegraphics[width=0.49\linewidth]{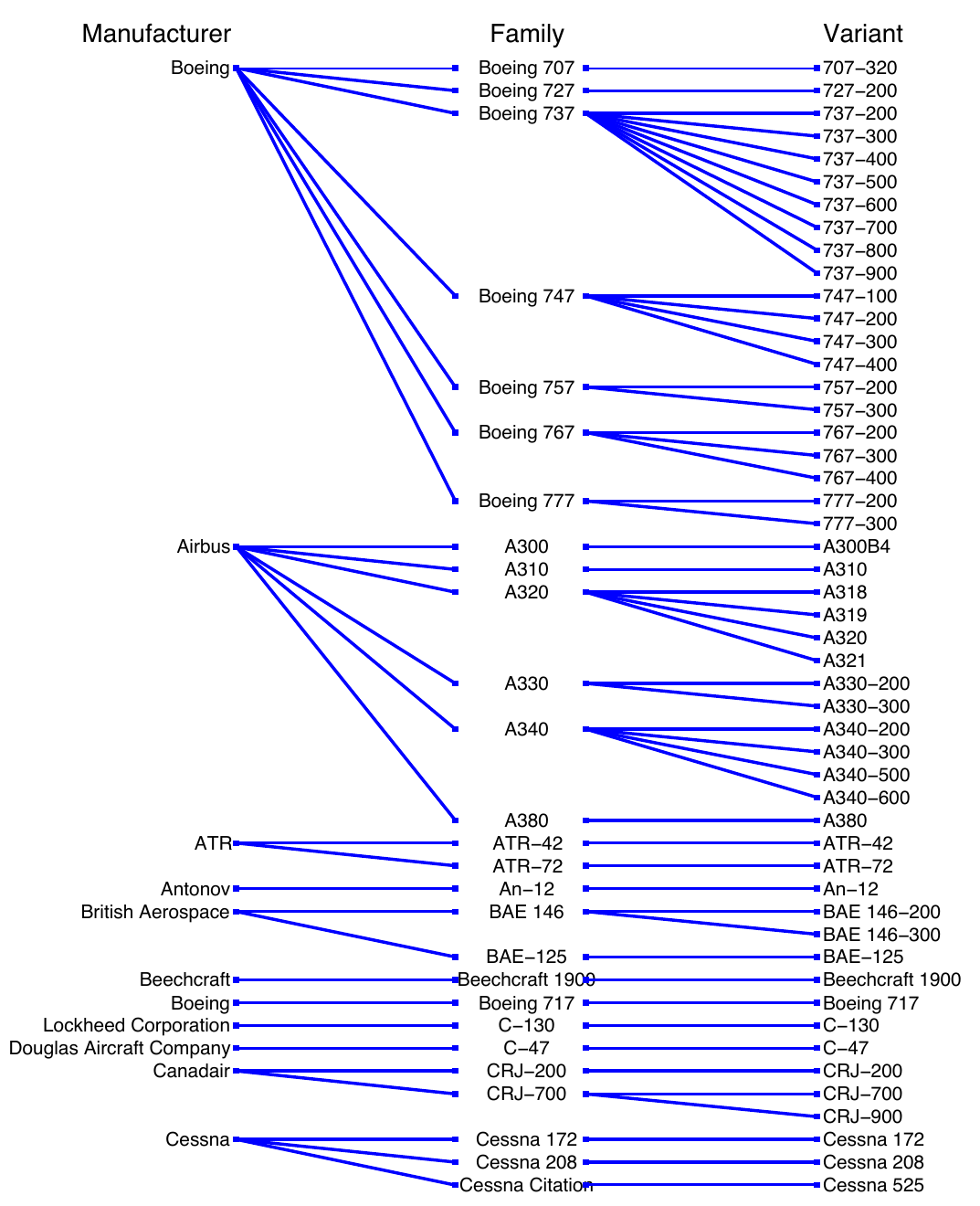} & 
\raisebox{0.13in}{\includegraphics[width=0.49\linewidth]{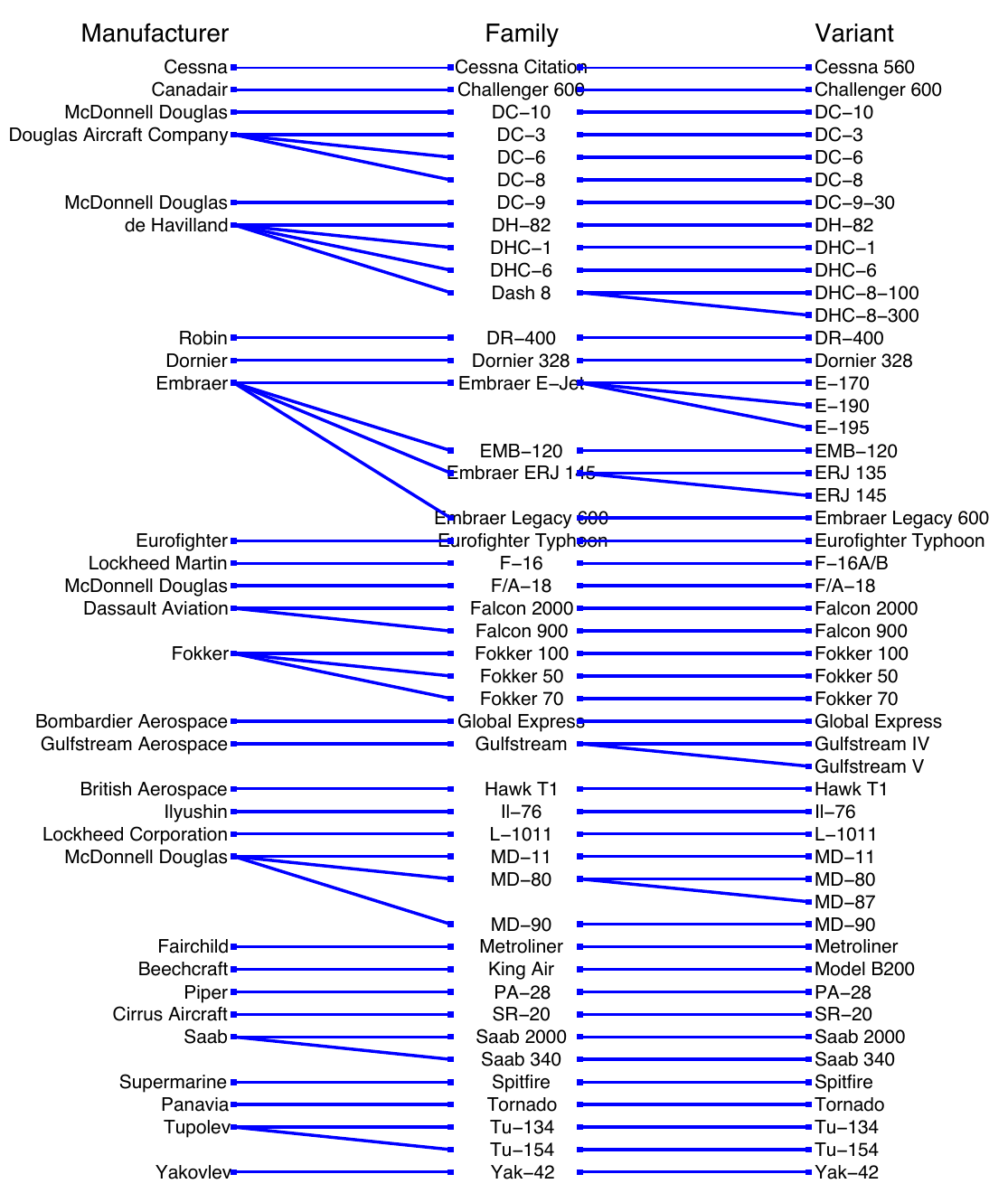}} \\
\end{tabular}
\caption{Label hierarchy shown as the \emph{manufacturer}, \emph{family} and the \emph{variant}. Our dataset contains aircrafts of 100 different variants grouped under 70 families and 30 manufacturers.}\label{f:labelHierarchy}
\end{figure*}

\begin{table*}
\centering
\begin{tabular}{|cc|cc|cc|}
\hline
\textbf{Model} & \textbf{Accuracy} & \textbf{Model} & \textbf{Accuracy} & \textbf{Model} & \textbf{Accuracy} \\
\hline
DR-400 & 94.1\%& DHC-8-100 & 57.6\%& ERJ 135 & 35.3\%\\ 
Eurofighter Typhoon & 94.1\%& Embraer Legacy 600 & 57.6\%& 747-100 & 33.3\%\\ 
F-16A/B & 90.9\%& F/A-18 & 57.6\%& 747-300 & 33.3\%\\ 
Cessna 172 & 88.2\%& 757-300 & 54.5\%& 767-200 & 33.3\%\\ 
SR-20 & 88.2\%& 767-400 & 54.5\%& 777-200 & 33.3\%\\ 
BAE-125 & 84.8\%& A340-500 & 54.5\%& BAE 146-200 & 33.3\%\\ 
DH-82 & 84.8\%& Cessna 208 & 54.5\%& DC-10 & 33.3\%\\ 
Tornado & 84.8\%& Challenger 600 & 54.5\%& DC-8 & 33.3\%\\ 
C-130 & 81.8\%& E-170 & 54.5\%& MD-87 & 33.3\%\\ 
Hawk T1 & 81.8\%& Gulfstream V & 54.5\%& 737-500 & 32.4\%\\ 
Model B200 & 81.8\%& ATR-42 & 51.5\%& 727-200 & 30.3\%\\ 
DHC-1 & 78.8\%& CRJ-900 & 51.5\%& A300B4 & 30.3\%\\ 
Il-76 & 76.5\%& EMB-120 & 51.5\%& A330-300 & 30.3\%\\ 
An-12 & 75.8\%& DC-3 & 50.0\%& E-190 & 29.4\%\\ 
Falcon 900 & 75.8\%& DHC-6 & 50.0\%& BAE 146-300 & 26.5\%\\ 
PA-28 & 75.8\%& Tu-134 & 48.5\%& 737-700 & 24.2\%\\ 
Spitfire & 70.6\%& Gulfstream IV & 47.1\%& A340-300 & 24.2\%\\ 
DC-6 & 69.7\%& Tu-154 & 47.1\%& MD-80 & 23.5\%\\ 
E-195 & 69.7\%& 737-900 & 45.5\%& A310 & 21.2\%\\ 
Cessna 560 & 67.6\%& Fokker 100 & 42.4\%& A319 & 21.2\%\\ 
Fokker 50 & 67.6\%& L-1011 & 42.4\%& A330-200 & 21.2\%\\ 
Cessna 525 & 66.7\%& Boeing 717 & 41.2\%& C-47 & 21.2\%\\ 
Global Express & 66.7\%& CRJ-200 & 41.2\%& 747-200 & 20.6\%\\ 
Saab 2000 & 66.7\%& DHC-8-300 & 39.4\%& 737-200 & 17.6\%\\ 
Yak-42 & 66.7\%& ERJ 145 & 39.4\%& 737-800 & 17.6\%\\ 
A318 & 64.7\%& ATR-72 & 38.2\%& 757-200 & 17.6\%\\ 
Falcon 2000 & 64.7\%& 707-320 & 36.4\%& A320 & 15.2\%\\ 
Metroliner & 64.7\%& 747-400 & 36.4\%& 767-300 & 14.7\%\\ 
Beechcraft 1900 & 63.6\%& CRJ-700 & 36.4\%& DC-9-30 & 14.7\%\\ 
Dornier 328 & 63.6\%& MD-11 & 36.4\%& 737-400 & 12.1\%\\ 
Fokker 70 & 63.6\%& MD-90 & 36.4\%& A321 & 11.8\%\\ 
Saab 340 & 63.6\%& 777-300 & 35.3\%& 737-300 & 06.1\%\\ 
737-600 & 57.6\%& A340-200 & 35.3\% & &\\ 
A380 & 57.6\%& A340-600 & 35.3\% & \textbf{Average} & \textbf{48.69\%}\\
\hline 
\end{tabular}
\caption{Accuracy of variant prediction sorted according to the accuracy for each of the 100 variants in our dataset.} \label{t:acc}
\end{table*}

% --------------------------------------------------------------------
\section{The dataset: content, tasks, and evaluation}\label{s:data}
% --------------------------------------------------------------------

FGVC-Aircraft contains 10,000 images of airplanes annotated with the model and bounding box of the dominant aircraft they contain. Aircraft models are organised in a four-level hierarchy, of which only the last three levels are of interest here.
\begin{itemize}
\item {\bf Model.} This is the most specific class label, such as {\em Boeing 737-76J}. This level is not considered meaningful for FGVC as differences between models may not be visually measurable, at least given an image of the exterior of the aircraft. \item {\bf Variant.}  Model variants are the finer distinction level that are visually detectable, and were obtained by merging visually indistinguishable models. For example, the variant {\em Boeing 737-700} includes 87 models such as {\em 737-7H4, 737-76N, 737-7K2, etc}. The dataset contains 100 variants.
\item {\bf Family.} Families group together model variants that differ in subtle ways, making differences between families more substantial. The goal of this level is to create a classification task of intermediate difficulty. For example, the family {\em Boeing 737} contains variants {\em 737-200, 737-300, \dots, 737-900}. The dataset contains 70 families.
\item {\bf Manufacturer.} A manufacturer is a grouping of families produced by the same company. For example, {\em Boeing} contains the families {\em 707, 727, 737, \dots}. The dataset contains airplanes made by 30 different manufacturers.
\end{itemize}
The list of model variants and corresponding example images are given in Fig.~\ref{f:variants} and the hierarchy is given in Fig.~\ref{f:labelHierarchy}.

FGVC-Aircraft contains 100 example images for each of the 100 model variants. The image resolution is about 1-2 Mpixels. Image quality varies as images were captured in a span of decades, but it is usually very good. The dominant aircraft is generally well centred, which helps focusing on fine-grained discrimination rather than object detection. Images are equally divided into training, validation, and test subsets, so that each  subset contains either 33 or 34 images for each variant. Algorithms should be designed on the training and validation subsets, and tested just once on the test subset to avoid over fitting.

Bounding box information can be used for training the aircraft classifiers, but should not be used for testing.

We define three tasks: aircraft variant recognition, aircraft family recognition, and aircraft manufacturer recognition. The performance is evaluated as class-normalised average classification accuracy, obtained as the average of the diagonal elements of the normalised confusion matrix. Formally, let $y_i \in \{1,\dots, M\}$ the ground truth label for image $i=1,\dots,N$ (where $N=10,000$ and $M=100$ for variant recognition). Let $\hat y_i$ be the label estimated automatically. The entry $C_{pq}$ of the {\em confusion matrix} is given by
\[
   C_{pq} = \frac{ | \{i : \hat y_i = q \wedge y_i = p\} |  }{ | \{ i : y_i = p \} |}
\]
where $| \cdot |$ denotes the cardinality of a set. The class-normalised average accuracy is then $\sum_{p=1}^M C_{pp}/M$.

The dataset is made publicly available {\em for research purposes only} at \dataseturl. Please note (Sect.~\ref{s:initial}) that the data contains images that were generously made available for research purposes by several photographers; however, these images should {\em not} be used for any other purpose without obtaining prior and explicit consent by the respective authors (see Sect.~\ref{s:ack} for further details). 

Authorship information is contained in a banner at the bottom of each image (20 pixels high).  {\em Do not forget to remove this banner before using the images in experiments.}

% --------------------------------------------------------------------
\section{Dataset construction}\label{s:construction}
% --------------------------------------------------------------------

Identifying the detailed model of an aircraft from an image is challenging for anyone but aircraft experts, and collecting 10,000 such annotations is daunting in general. Sect.~\ref{s:initial} explains how leveraging aircraft data collected by {\em aircraft spotters} was the key in the construction of FGVC-Aircraft. However, collecting data from a restricted number of sources presents its own challenges. Sect.~\ref{s:diversity} introduces a notion of diversity and applies it to select a subset of the data that is maximally uncorrelated. Sect.~\ref{s:finish} explains how bounding box annotations were crowdsourced using Amazon Mechanical Turk, and Sect.~\ref{s:hierarchy} how the hierarchical labels were obtained.

% --------------------------------------------------------------------
\subsection{Initial data collection}\label{s:initial}
% --------------------------------------------------------------------

Enthusiasts, collectors, and other hobbyists may be an excellent source of annotated visual data. In particular, data obtained from aircraft spotters was instrumental in the construction of this FGVC-Aircraft. A large number of such annotated images is available online in Airliners.net (\url{http://www.airliners.net/}), a repository of aircraft spotting data (similar collections exists, for example, for cars and trains). While using such images for research purposes may be considered {\em fair use}, nevertheless we found appropriate to ask for explicit permission to the photographers due to the large quantity of data involved. Of about twenty photographers that were contacted, permission to use the data for research purposes was granted by about ten of them (Sect.~\ref{s:ack}), and an explicit negative answer was received only from two of them. FGVC-Aircraft contains data only from the photographers that  explicitly made their pictures available (see Sect.~\ref{s:data} and Sect.~\ref{s:ack} for further details).

About 70,000 images were downloaded from the ten photographers, resulting in images spanning thousands of different aircraft models. Even after grouping these models into variants, there was still a very large number of different classes, with a very skewed distribution. Popular families such as Airbus and Boeing included thousand of images per model variant, whereas rarer models counted only a  dozen images. The 100 most frequent variants
were retained, resulting in at least 120 images per variant.

% --------------------------------------------------------------------
\subsection{Diversity maximisation}\label{s:diversity}
% --------------------------------------------------------------------

One drawback of relying on a small set of photographers is that unwanted correlation may be introduced in the data. While these photographers tend to be active in the span of several years, it is natural to expect at least regional dependencies (for example certain airliners may fly more frequently to certain airports). Therefore, the data was first filtered to maximise internal diversity. Each pair of images for a given variant was compared based on photographer, time, airliner, and airport, obtaining an ``a priori'' similarity score (i.e., without looking at the pictures). Then, 100 images per variant were incrementally and greedily selected in order of decreasing diversity to the images already added to the collection. After doing so, images were randomly assigned to the training, validation, and test subsets. This simple procedure was effective at reducing internal correlation in the data, as reflected by a substantial reduction of the classification performance of baseline classifiers. In particular, sequences of photos are broken whenever possible.

Isolating different photographers in different splits was also considered as an option, but ultimately it was rejected due to the complex dependency structure that such a choice would have introduced in the data.

% --------------------------------------------------------------------
\subsection{Bounding boxes}\label{s:finish}
% --------------------------------------------------------------------

About 110 images were initially selected for each variant and submitted to Amazon Mechanical Turk for bounding box annotation. Annotators were instructed to skip images that did not contain the exterior of an aircraft, so that these images could be identified and discarded. Three annotations were collected for each image, presenting annotators with batches of 10 images at a time and paying 0.03 USD per batch. Overall, the cost of annotating all the images was 110 USD and annotations were complete in less than 48 hours. Out of three annotations, we sought at least two whose overlap over union similarity score was above 0.85\% (fairly restrictive in practice), discarding other annotations. The remaining annotations were then averaged to obtain the final bounding box, and images without a bounding box (usually due to a problematic image) were discarded. Since slightly more than 100 images were submitted for annotation, this eventually resulted in a sufficient number of validated images.

% --------------------------------------------------------------------
\subsection{Hierarchy}\label{s:hierarchy}
% --------------------------------------------------------------------

The hierarchy (Fig.~\ref{f:labelHierarchy}) was obtained largely by manual inspection. Fortunately, sorting models by name is very likely to suggest possible merges in a straightforward way. These were verified manually by searching example images, Wikipedia, and the manufacturer websites for clear evidence that two model would differ visually. If no evidence was found, then the two models were merged in a variant. Sometimes, differences are fairly subtle; for example, Boeing variants -200, -300, -400, \dots  differ mostly in length, an attribute that is difficult to estimate from monocular images (in this case counting windows may be the best way of telling a model from another).

% --------------------------------------------------------------------
\section{Baselines}\label{s:baselines}
% --------------------------------------------------------------------

We consider the classification tasks given in Sect.~\ref{s:data}. For example, the variant classification for our dataset is a 100-way binary classification problem and performance is measured in term of class-normalised average accuracy as described earlier.

\begin{figure*}
\centering
\begin{tabular}{rl}
\includegraphics[width=0.9\linewidth]{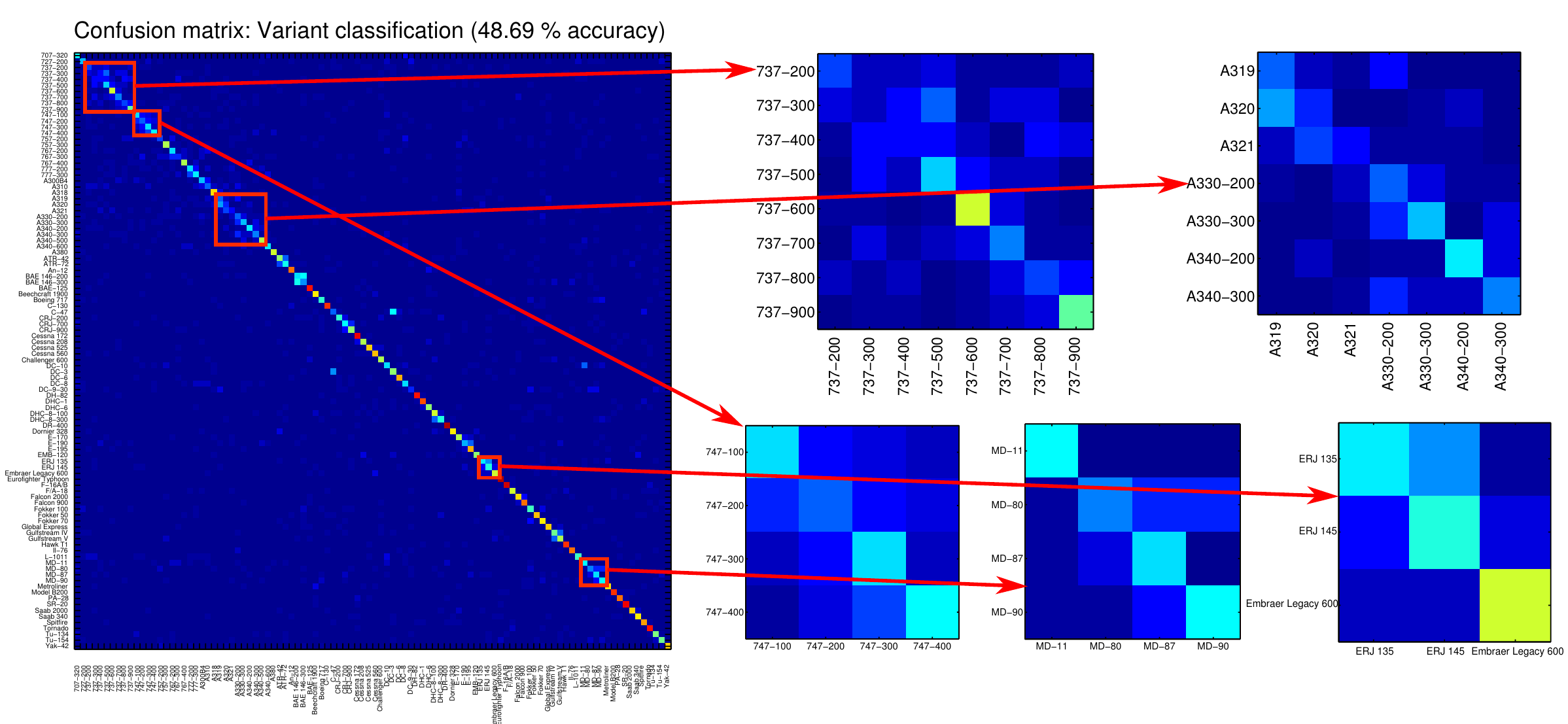} & 
\includegraphics[width=0.0745\linewidth]{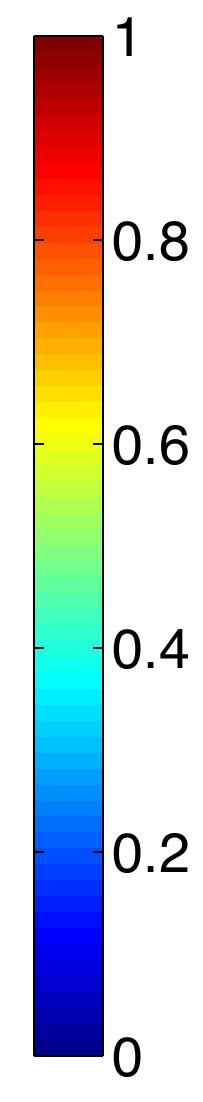} \\
\end{tabular}
\caption{Confusion matrix for the 100 variant classification challenge. Some high confusion, due to the similarity of the models are also shown. These correspond to the Boeing 737 family, Boeing 747 family, Airbus family, McDonnell Douglas (MD) and the Embraer family. The average diagonal accuracy is 48.69\%.}\label{f:conf}
\end{figure*}

Fig.~\ref{f:conf} shows the confusion matrix for a strong baseline model (non-linear SVM on a $\chi^2$ kernel, bag-of-visual words, 600 k-means words dictionary, multi-scale dense SIFT features, and $1\times 1$, $2\times 2$ spatial pyramid~\cite{chatfield11the-devil}). These models were trained on the entire image ignoring the bounding box information. As seen in Tab.~\ref{t:acc} the performance is quite good for a few relatively distinctive categories (e.g., the ``Eurofighter Typhoon'' has error of just 5.9\%). On the other hand, bag-of-visual-words is much worse at picking up subtle variations, such as for Airbus or Boeing family, resulting in large intra-family confusion (Fig.~\ref{f:conf}). The overall accuracy of the classifier is 48.69\%.

\begin{figure*}
\centering
\begin{tabular}{ccl}
\includegraphics[width=0.4375\linewidth]{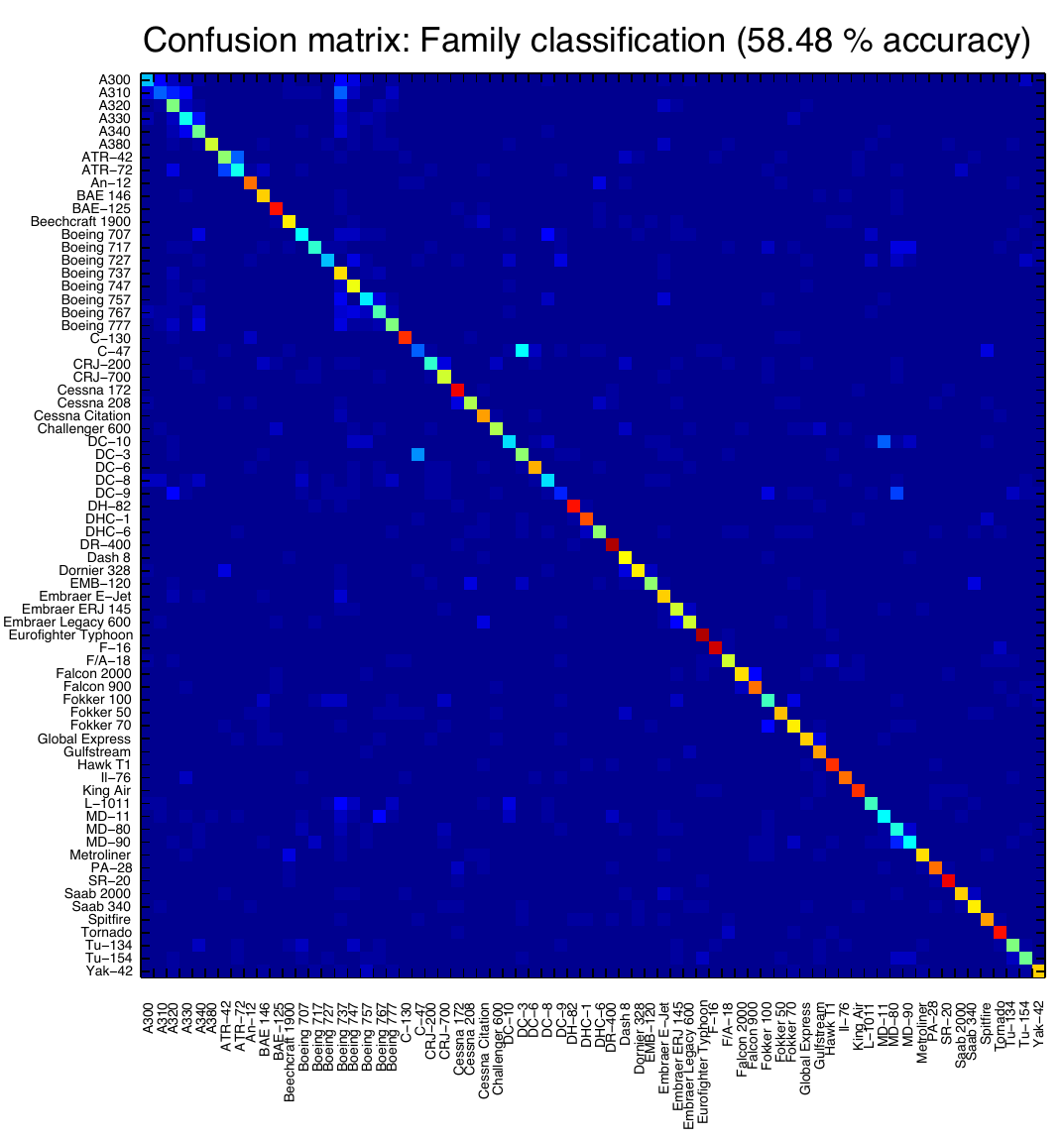} & 
\includegraphics[width=0.4375\linewidth]{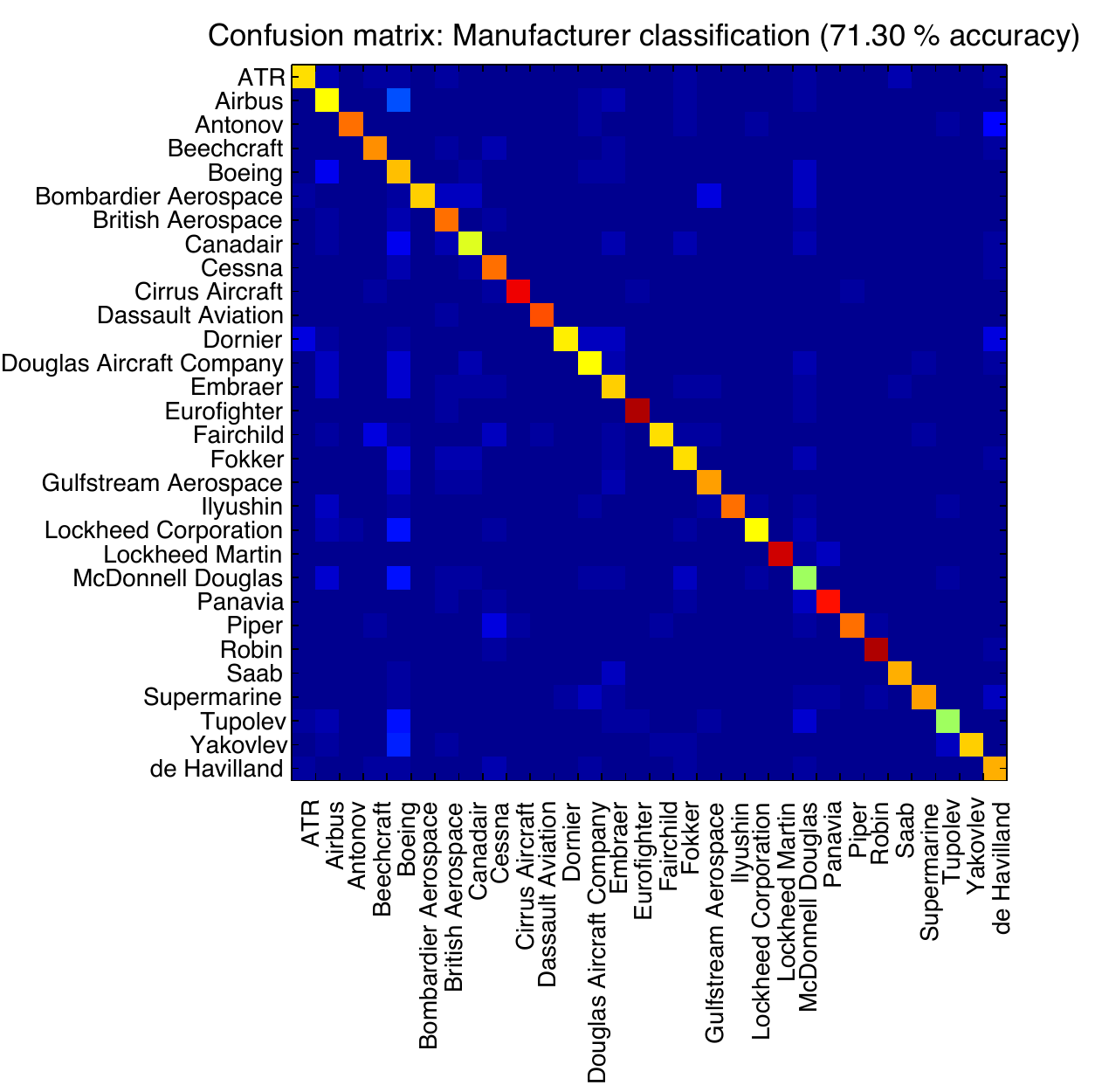}  &
\includegraphics[width=0.0745\linewidth]{f/baselineConfusion/colorbar.pdf} \\

\end{tabular}
\caption{Confusion matrix for the family (\emph{left}) and  manufacturer (\emph{right}) classification tasks.}\label{f:confHierarchy}
\end{figure*}

Fig.~\ref{f:confHierarchy} shows the accuracy of the classifier when measured on the hierarchical label classification tasks. The accuracy for the variant classification is  58.48\%, whereas, the accuracy for manufacturer classification is 71.30\%. At the top level the two manufacturers, Boeing and Airbus, are most confused with one another perhaps due to the similar kinds of aircrafts they manufacture -- large passenger planes catering to airliners. Note that for the hierarchical evaluation we trained our models for the variant classification task and simply measured the performance at different levels of the hierarchy by merging the labels below. An alternative strategy, which is to train the models directly for the labels at a given level in the hierarchy, performed significantly worse in our experiments. 

% --------------------------------------------------------------------
\section{Summary}\label{s:summary}
% --------------------------------------------------------------------

We have introduced FGVC-Aircraft, a new large dataset of aircraft images for fine-grained visual categorisation. The data contains 10,000 images, 100 airplane model variants, 70 families, and 30 manufacturers. We believe that FGVC-Aircraft has the potential of introducing aircraft recognition as a novel domain in FGVC to the wider computer vision community (FGVC-Aircraft will be part of the ImageNet 2013 FGVC challenge). Compared to other classes used frequently in FGVC, aircraft have different and interesting modes of variation.

Images in FGVC-Aircraft were obtained from aircraft spotter collections, maximising internal diversity in order to reduce unwanted correlation between images taken by a limited number of photographers; in the future, we plan to substantially increase the size of the FGVC-Aircraft dataset by including more models as more and more photographers provide permission to use their photos, and apply the same construction to other object categories as well.

% --------------------------------------------------------------------
\subsection{Acknowledgments}\label{s:ack}
% --------------------------------------------------------------------

The creation of this dataset started during the Johns Hopkins CLSP Summer Workshop 2012, \emph{Towards a Detailed Understanding of Objects and Scenes in Natural Images}\footnote{\small{\url{http://www.clsp.jhu.edu/workshops/archive/ws-12/groups/tduosn/}}.} with, in alphabetical order, \emph{Matthew B. Blaschko, Ross B. Girshick, Juho Kannala, Iasonas Kokkinos, Siddharth Mahendran, Subhransu Maji, Sammy Mohamed, Esa Rahtu, Naomi Saphra, Karen Simonyan, Ben Taskar, Andrea Vedaldi, and David Weiss}. The CLSP workshop was supported by the National Science Foundation via Grant No 1005411, the Office of the Director of National Intelligence via the JHU Human Language Technology Center of Excellence; and Google Inc. A special thanks goes to \emph{Pekka Rantalankila} for helping with the creation of the airplane hierarchy.

Many thanks to the photographers that kindly made available their images for research purposes. These are, in alphabetical order,
\emph{Mick Bajcar,
Aldo Bidini,
Wim Callaert,
Tommy Desmet,
Thomas Posch,
James Richard Covington,
Gerry Stegmeier,
Ben Wang,
Darren Wilson} and \emph{
Konstantin von Wedelstaedt}. \emph{Please note that images are made available exclusively for non-commercial research purposes. The original authors retain the copyright on the respective pictures and should be contacted for any other usage of them.  Photographers may be contacted through their \url{http://www.airliners.net} profile pages, which are linked from \dataseturl.}

% --------------------------------------------------------------------
{\footnotesize
  \bibliographystyle{plain}
  \bibliography{fgvc}
}
\end{document}